# Stroke Lesions as a Rosetta Stone for Language Model Interpretability

Julius Fridriksson[1,2*], Roger D. Newman-Norlund[1,2], Saeed Ahmadi[1], Regan Willis[3], Nadra Salman[4], Kalil Warren[4], Xiang Guan[3], Yong Yang[3], Srihari Nelakuditi[3], Rutvik Desai[5], Leonardo Bonilha[6], Jeff Charney[2,7], Chris Rorden[5]

[1] *Department of Communication Sciences and Disorders, University of South Carolina*

[2] *ALLT.AI, LLC, Columbia, SC*

[3] *Department of Computer Science and Engineering, University of South Carolina*

[4] *Linguistics Program, University of South Carolina*

[5] *Department of Psychology, University of South Carolina*

[6] *Department of Neurology, USC School of Medicine*

[7] *MKHSTRY, LLC, Cleveland, OH*

*Corresponding author: fridriks@mailbox.sc.edu

## Abstract

*Large language models (LLMs) have achieved remarkable capabilities, yet methods to verify which model components are truly necessary for language function remain limited. Current interpretability approaches rely on internal metrics and lack external validation. Here we present the Brain-LLM Unified Model (BLUM), a framework that leverages lesion-symptom mapping—the gold standard for establishing causal brain-behavior relationships for over a century—as an external reference structure for evaluating LLM perturbation effects. Using data from individuals with chronic post-stroke aphasia (N = 410), we trained symptom-to-lesion models that predict brain damage location from behavioral error profiles, applied systematic perturbations to transformer layers, administered identical clinical assessments to perturbed LLMs and human patients, and projected LLM error profiles into human lesion space. LLM error profiles were sufficiently similar to human error profiles that predicted lesions corresponded to actual lesions in error-matched humans above chance in 67% of picture naming conditions (p < 10 ⁻²³) and 68.3% of sentence completion conditions (p < 10 ⁻⁶¹), with semantic-dominant errors mapping onto ventral-stream lesion patterns and phonemic-dominant errors onto dorsal-stream patterns. These findings open a new methodological avenue for LLM interpretability in which clinical neuroscience provides external validation, establishing human lesion-symptom mapping as a reference framework for evaluating artificial language systems and motivating direct investigation of whether behavioral alignment reflects shared computational principles.*

# 1. Introduction

## 1.1 The Validation Gap in Large Language Model Interpretability

Large language models (LLMs) have achieved remarkable capabilities across diverse linguistic and cognitive tasks, yet a fundamental challenge persists: methods to verify which model components are truly necessary for language function lack external grounding. Perturbation-based approaches—including pruning, layer ablation, and causal tracing—can identify components whose removal degrades performance on specific benchmarks (Men et al. 2025; Gromov et al. 2025; Meng et al. 2022; Hase et al. 2023). However, these benchmarks were designed primarily to assess reasoning, knowledge retrieval, and general cognitive performance rather than to isolate foundational language processing. They reflect considered judgments about what constitutes useful AI capability, but they are not grounded in principled accounts of human language or cognition. Without such grounding, interpretability research can identify components necessary for benchmark performance, but cannot establish whether those components implement anything fundamental about language itself.

This limitation reflects a deeper absence of ground truth. Probing classifiers may reveal that syntactic information is recoverable from specific activations, but this does not establish whether such information is organized as it is in systems that actually process language (Belinkov 2022; Hewitt and Manning 2019; Conneau et al. 2018; Belinkov and Glass 2019; Ravichander et al. 2021). Magnitude-based pruning may remove parameters without degrading benchmark performance, but benchmarks capture only what their designers chose to measure (Mohanty et al. 2024). Sparse autoencoders may discover features that support accurate reconstruction, but whether these features carve language at its joints—or merely at joints convenient for reconstruction—cannot be determined through internal metrics (Shu et al. 2025; Templeton et al. 2024; Cunningham et al. 2023). The limitations of these metrics are increasingly evident: models can maintain performance on benchmarks such as MMLU even when 40% of their layers are removed, while next-token prediction on held-out data degrades proportionally to pruning depth (Gromov et al. 2025; Men et al. 2025). If a model can lose nearly half its computational depth without degrading on a "language understanding" benchmark, either the pruned layers contributed nothing to language understanding, or the benchmark does not measure it. Without external grounding, there is no principled way to adjudicate. The proliferation of benchmarks does not resolve this problem; it multiplies evaluation metrics without establishing contact with the phenomenon being modeled.

LLMs are, fundamentally, models of human language. Language is not an abstract formal system but a biological capacity that emerges from human neural organization. The only systems that genuinely produce and comprehend language—as opposed to generating statistically plausible approximations—are human brains. This makes human language processing not merely one possible reference point among many, but the appropriate ground truth against which language models can be evaluated.

Moreover, the human language system did not evolve to optimize benchmark performance or token prediction accuracy. Evolution solves for survival, not for any particular cognitive capacity in isolation (Buzsáki 2019). Language probably emerged in service of survival, shaped by pressures fundamentally different from those governing LLM training—pressures involving real-time communication, social coordination, and adaptive behavior in uncertain environments. The brain's organization of language reflects these evolutionary constraints: robust, efficient under severe metabolic limitations, and integrated with perception, action, and social cognition. This makes convergence between artificial and biological language systems a meaningful empirical question rather than an expected outcome. Where LLM perturbation produces breakdown patterns that correspond to human lesion-symptom relationships, we gain evidence that the artificial system has captured aspects of how language is organized in biological systems. Where correspondence fails, we learn that the model—whatever computations it performs—is solving a different problem than the one evolution solved.

If human language processing provides the appropriate ground truth for evaluating LLMs, then the study of language breakdown in humans becomes directly relevant to understanding artificial language systems. Aphasia—the impairment of language following brain damage—has yielded over a century of causal evidence about how the human language system is organized (Broca 1861; Wernicke 1874; Geschwind 1970). Unlike correlational neuroimaging methods, lesion studies establish which neural substrates are necessary for specific language functions: when damage to a region consistently produces a specific deficit, that region is causally implicated in the corresponding computation. Decades of research in individuals with post-stroke aphasia have mapped the functional architecture of human language with remarkable precision, identifying dissociable neural substrates for phonological, semantic, syntactic, and fluency processes (Fridriksson et al., 2018; Mirman et al., 2015; Bates et al., 2003; Dronkers et al., 2004). Modern multivariate approaches account for more than 60% of the variance in chronic aphasia severity from lesion location alone (Pustina et al., 2017; DeMarco and Turkeltaub, 2018). This methodology provides exactly what LLM interpretability lacks—an externally grounded, causally validated map of how a real language system is organized.

Comparing how LLMs fail under perturbation to how humans fail following brain damage is therefore not an arbitrary exercise in cross-system comparison. It is the principled application of the only available ground truth for language processing to the evaluation of systems designed to model language. Crucially, this approach respects our ignorance: we do not need a complete theory of human cognition to learn from empirical patterns of language breakdown. By using data accumulated over 150 years of clinical observation, we can let the evidence reveal which aspects of LLM organization correspond to biological necessity and which do not, without presupposing that we understand either system completely. If perturbing an LLM component produces errors that correspond to those produced by damage to specific brain regions, we gain evidence that the component implements computations related to those performed by the neural substrate. If perturbation produces errors that bear no resemblance to any human deficit pattern, we conclude that the component—whatever function it serves—does not implement language processing as it occurs in biological systems.

## 1.2 Lesion–Symptom Mapping as External Ground Truth

A central strength of lesion-based approaches is their ability to identify neural regions that are indispensable for successful task performance. Whereas popular activation-based measures indicate which regions are engaged during a task, lesions reveal which computations are truly necessary. Functional neuroimaging and brain-model alignment approaches largely rely on correlational associations, which cannot distinguish whether a region or representation is necessary for a given computation. In contrast, focal brain lesions offer direct evidence of causal necessity by revealing which computations fail when specific neural substrates are destroyed. Lesion–symptom mapping has served as the gold standard for establishing causal brain-behavior relationships for well over a century (Broca 1861; Wernicke 1874; Geschwind 1970). Stroke provides natural experiments revealing which computations cannot be adequately compensated by remaining tissue and which can be lost with minimal consequence. Decades of research in individuals with post-stroke aphasia have established that the human language system comprises distributed but functionally specialized regions whose contributions can be causally identified (Bates et al. 2003; Dronkers et al. 2004; Fridriksson et al. 2018). Modern multivariate approaches predict over 60% of variance in chronic aphasia severity from lesion location alone (Pustina et al. 2017; DeMarco and Turkeltaub 2018), and large-scale studies consistently identify dissociable neural substrates for phonological, semantic, syntactic, and fluency processes (Fridriksson et al. 2018; Mirman et al. 2015). This body of work provides precisely what LLM interpretability currently lacks: an external criterion for distinguishing essential from dispensable computations.

## 1.3 The Brain–LLM Unified Model Framework

Building on this motivation, we introduce the Brain–LLM Unified Model (BLUM), a framework that uses human lesion–symptom relationships as an external reference structure for interpreting large language models (LLMs) (Figure 1). BLUM does not assume that biological and artificial systems implement language through identical mechanisms, nor does it necessarily treat the human brain as a template that LLMs are expected to replicate. Instead, it leverages a well-established causal map of human language organization to ask a more constrained and falsifiable question: whether the error patterns produced by perturbed LLMs can be meaningfully situated within a neurobiologically grounded space defined by human functional necessity. In humans, decades of lesion–symptom mapping demonstrate that patterns of spoken errors carry information about which neural substrates are causally indispensable for language performance. BLUM takes this causal structure as given on the human side and uses it as an external standard against which LLM behavior can be evaluated.

The core premise of BLUM is that behavioral error profiles can serve as a shared representational interface between humans and language models, even when the underlying substrates and learning histories differ radically. Symptom-to-lesion models are trained exclusively on human data, where irreversible neural damage licenses causal inference and establishes a mapping from error structure to neuroanatomical necessity. Error profiles generated by perturbed LLMs are then projected into this human-defined lesion space. If LLM error patterns were arbitrary, idiosyncratic, or organized along dimensions unrelated to human language function, this projection would either fail or yield anatomically nonspecific predictions. Systematic alignment, by contrast, would indicate that the dimensions governing breakdown in LLM performance under perturbation overlap with dimensions of breakdown that are causally critical in the human brain. Importantly, such alignment does not imply mechanistic equivalence, biological homology, or independent rediscovery of neuroanatomical organization by the model; rather, it demonstrates that LLM behavior can be embedded within a space defined by human causal constraints.

The BLUM pipeline proceeds in three stages. First, symptom-to-lesion models are trained on human aphasia data to predict lesion location from behavioral error profiles alone, establishing that error structure contains sufficient information to reconstruct neuroanatomical damage. Second, systematic perturbations are applied to LLM components, and the same clinically validated language assessments used with human patients are administered to the perturbed models, with outputs classified using an identical error taxonomy. Third, LLM error profiles are projected into human lesion space using the human-trained models, and the resulting predicted lesions are evaluated by comparison to actual lesions observed in humans with similar error profiles. If lesion patterns inferred from LLM error profiles correspond to lesion patterns observed in behaviorally matched humans more strongly than expected by chance, this provides evidence that LLM error behavior reflects functionally meaningful structure aligned with human language organization. In this way, BLUM establishes human clinical data as an external, causally grounded reference point for evaluating which perturbations and components of LLMs are functionally consequential, complementing internal-model interpretability methods with an independent standard rooted in human neuroscience.

## 1.4 Present Study

The present study introduces BLUM as an initial experimental test of a broader framework for using human lesion–symptom relationships as external ground truth for LLM interpretability. Using data from individuals with chronic post-stroke aphasia and a 13-billion-parameter transformer-based language model, we apply BLUM to two well-characterized, complementary language tasks: picture naming (Philadelphia Naming Test: PNT) and sentence completion (Western Aphasia Battery–Revised: WAB-R). We selected picture naming because anomia is the hallmark impairment in aphasia, present across all subtypes and severity levels (Goodglass and Kaplan 1983; Laine and Martin 2023). We selected sentence completion because it mirrors the next-word prediction task, which is fundamental to LLM architecture. By administering identical assessments to both human patients and perturbed LLMs and classifying responses using a common error taxonomy, we evaluated whether LLM-derived predicted lesions correspond to actual lesions in error-matched humans, and whether these predictions respect the anatomical organization established by decades of human lesion–symptom mapping. Importantly, the present analyses are intentionally limited in scope. The goal here is not to exhaustively model human language, but to establish proof of concept: to test whether even a restricted set of clinically grounded behavioral dimensions is sufficient to reveal nontrivial, anatomically meaningful correspondence between perturbed LLMs and the aphasic human brain. This is a necessary first step towards extending BLUM to richer linguistic features and more sophisticated architectures in future work.

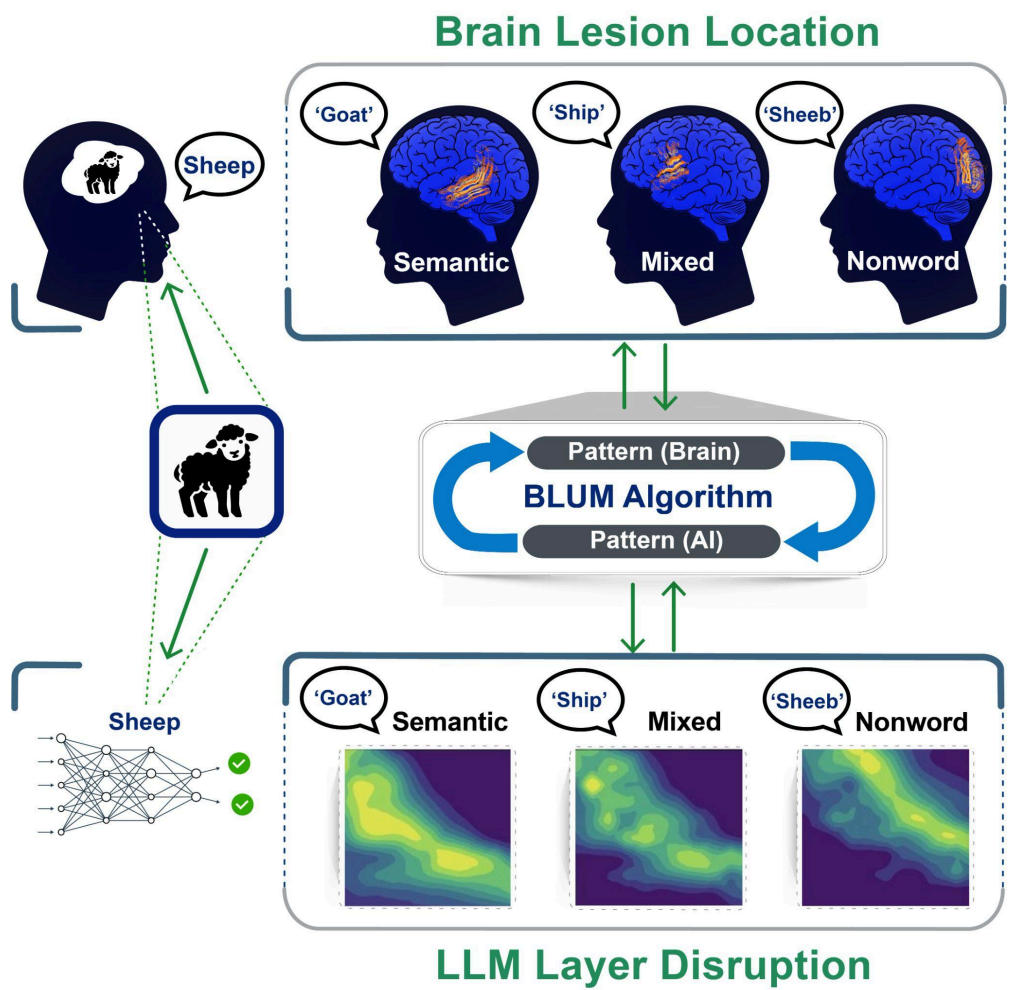

**Figure 1. The Brain-LLM Unified Model (BLUM) framework.** The BLUM approach establishes systematic mappings between lesion–deficit relationships in the human brain and perturbation–deficit relationships in large language models (LLMs). Left: Both humans with aphasia (top) and LLMs (bottom) perform a picture-naming task. Different error types emerge from each system, including semantic errors ('Goat'), mixed errors ('Ship'), and phonological errors ('Sheeb'). Top right: In humans, distinct error types are associated with damage to different brain regions. Bottom right: Different error types in LLMs are associated with perturbations to different transformer layers. Center: The BLUM algorithm maps LLM error profiles into human brain space using symptom-to-lesion models, enabling validation against actual human lesions.

## 2. Methods

### 2.1 Overview

The BLUM pipeline proceeds in three stages (Figure 2). First, we establish symptom-to-lesion models in humans that predict lesion location solely from spoken error profiles, validating that behavioral patterns contain sufficient neuroanatomical information to predict lesion patterns. Second, in the LLM, we apply systematic perturbations to transformer layers, varying target layer, percent disruption, and noise level, and classify the resulting output errors using the same taxonomy applied to human behavior to characterize error distributions across perturbation conditions. Third, we map LLM error profiles into human brain space by passing them through our human data-based symptom-to-lesion models and validate the resulting predicted lesions against actual lesions in humans with similar error profiles.

This design tests whether LLM error patterns contain information that maps onto human neuroanatomy in predictable ways. Again, correspondence between LLM-derived predictions and actual human lesions would validate BLUM as a framework for using clinical data as external ground truth for LLM interpretability.

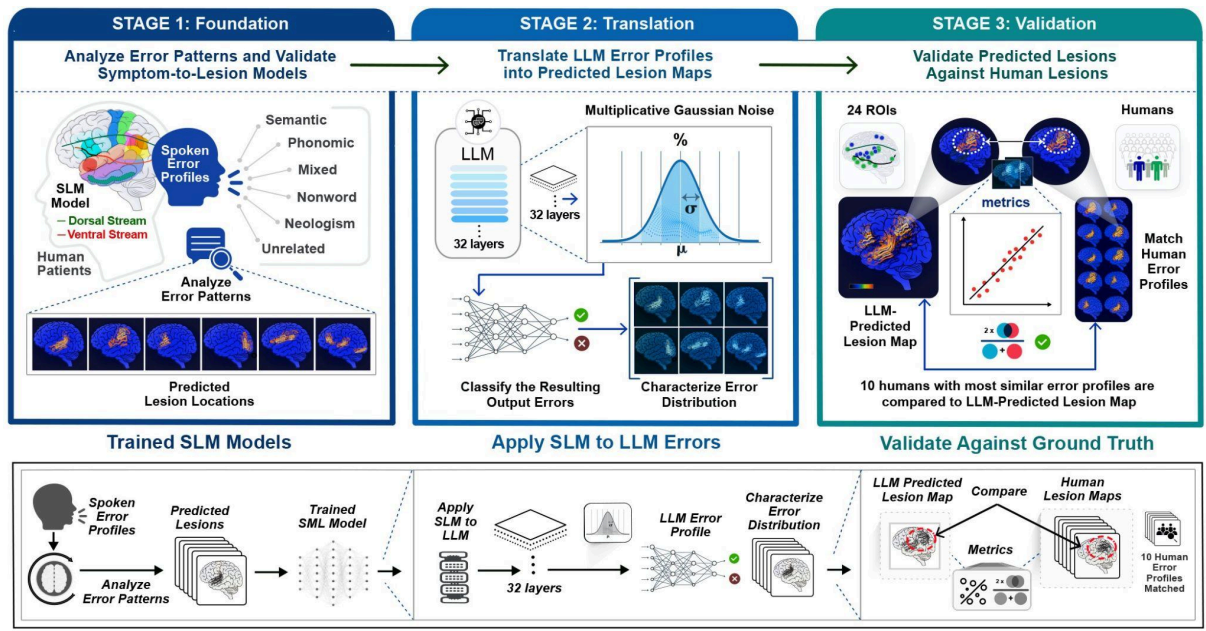

**Figure 2. BLUM pipeline.** First, we applied our patented symptom-to-lesion pipeline to build predictive models that map aphasia error patterns to lesion locations in human patients. Next, we input error profiles generated by large language models (LLMs) into these human-trained models to generate corresponding *virtual* brain lesions. We then identified the five human patients whose aphasia error profiles most closely matched each LLM error profile and computed the average of their lesion maps. To assess alignment, we calculated the Pearson correlation between this averaged human lesion and the LLM-predicted lesion and compared it, using permutation testing (2,000 permutations),

to correlations obtained between the LLM-predicted lesion and averaged lesions from five randomly selected human patients. This analysis demonstrates that LLM error patterns encode neurobiologically meaningful information.

## 2.2 Human Foundation Data

Data were drawn from an archival database maintained by the Aphasia Lab at the University of South Carolina. Although the original database contains data from 410 patients with chronic post-stroke aphasia, the final sample used in the analyses presented here was 214 (PNT) and 184 (WAB-R). This reduction was caused by not all participants having individual spoken errors transcribed on the aphasia tests or insufficient lesion data. Despite the reduced sample size, the final sample represents one of the largest lesion-symptom mapping studies to date. All participants had left-hemisphere-only stroke confirmed by neuroimaging, presence of aphasia as determined by clinical evaluation, and premorbid proficiency in English.

Participants completed the Western Aphasia Battery–Revised (WAB-R) and the Philadelphia Naming Test (PNT). We developed a unified error classification system to categorize responses as either correct or as one of six error types: Correct, Semantic, Phonemic (combining Formal and Nonword errors), Mixed, Neologism, No Response, and Unrelated. The automated classification system achieved >97% agreement with consensus human annotations (see Supplementary Methods S1 for the full decision tree and inter-rater reliability analysis).

Structural MRI data were acquired for all participants. Lesion masks were manually demarcated on T2-weighted images and normalized to Montreal Neurological Institute space using enantiomorphic normalization (see Supplementary Methods S2 for neuroimaging details). Normalized lesion maps were intersected with the Johns Hopkins University (JHU) atlas, yielding lesion load values (proportion of voxels damaged) for 189 cortical and white matter regions. Lesion coverage across participants is shown in Supplementary Figure S1.

## 2.3 Symptom-to-Lesion Models

We trained models to predict lesion location solely from error profiles, laying the foundation for mapping LLM errors into brain space. For each participant, we paired their error profile (proportions of each error type across WAB-R and PNT responses) with their lesion map (proportion of damage for each of 24 a priori language-related cortical regions). These areas constituted a subset of the 189 regions of interest (ROI) in the Johns Hopkins University (JHU) neuroanatomical atlas (Faria et al. 2012) and were based on their empirically demonstrated relationship to language production and comprehension (Fridriksson et al. 2018; Papagno 2011; Janelle et al. 2022; Nogueira et al. 2025). Details of the lesion-symptom mapping statistical framework are provided in Supplementary Methods S3.

We employed a per-ROI regression framework: for each of the 24 a priori left-hemisphere ROIs, we trained a linear regression model predicting lesion load from error proportions. Model performance was evaluated using leave-one-out cross-validation, computing $R^2$ and Pearson correlation between predicted and actual lesion loads for held-out participants. A final model, trained on all participant data, was saved for future stages of the pipeline.

This approach tested whether error profiles contain sufficient information to predict lesion location. Successful prediction validates the symptom-to-lesion direction and establishes trained models that can be applied to any error profile, including those generated by perturbed LLMs.

### 2.4 LLM Selection and Lesioning

For all LLM analyses, we used a Large Language and Vision Assistant (LLaVA-1.6-Vicuna-13B) model. This model has 13 billion parameters and is a state-of-the-art vision–language model (Liu et al. 2023). Our choice to use this model was dictated by methodological constraints of the BLUM framework. Picture naming (PNT) requires direct visual input and single-word responses, which precludes text-only language models without introducing external vision pipelines that would confound interpretation; LLaVA enables end-to-end image-to-word processing within a unified architecture. All perturbations and analyses (for both PNT and WAB-R-related processing) utilized the Vicuna language model component, ensuring comparability across tasks. Vicuna-13B provides sufficient scale to exhibit graded, human-like error profiles under perturbation while remaining fully accessible for systematic, layer-wise causal intervention, unlike larger proprietary models (Chiang et al. 2023). Our aim was not to compare architectures but to test whether a contemporary, perturbable transformer exhibits error structures that can be meaningfully projected into human lesion space.

We evaluated the model on WAB-R sentence-completion and responsive-speech items, the same items administered to human participants. Outputs were generated with deterministic decoding (do_sample=False) and fixed random seeds to ensure reproducibility. LLM outputs were classified using the same error taxonomy applied to human responses.

Perturbations were applied to individual transformer layers using multiplicative Gaussian noise. For each weight $w$ in a targeted layer, we sampled a noise term $\varepsilon \sim \mathcal{N}(0, \sigma^2)$, where $\varepsilon$ denotes a zero-mean random perturbation and $\sigma$ controls its standard deviation (i.e., perturbation severity), and updated the weight as $w' = w \times (1 + \varepsilon)$. Each perturbation condition was defined by: layer index (1–40), modification percentage (proportion of weights perturbed 10–19%, in 10% increments), and noise standard deviation ($\sigma$=0.1–, in 0.1-unit increments). For each given condition of layer, modification percentage, and noise, the subset of nodes selected for perturbation was randomly chosen from the 5,120 hidden units in the targeted layer. This parameterization allows systematic exploration of lesion location (layer), extent (modification percentage), and severity (noise level). Additional implementation details and parameter sweeps are described in Supplementary Methods S4.

## 2.5 Mapping LLM Errors to Brain Space

For each LLM perturbation condition, we first reduced the dataset by removing degenerate error profiles (total error mass < 0.02, effectively error-free; the No Response category exceeding 0.95, indicating model collapse to nonresponse; or entropy <0.4 bits, indicating error mass concentrated in a single category), resulting in ~1,987 nondegenerate PNT LLM conditions and 2,228 nondegenerate WAB-R LLM conditions. We then extracted the error profile (proportions of each error type) from each LLM condition and passed it to our human data-based symptom-to-lesion model to generate predicted lesion load for each of the 24 language-related ROIs, constituting a predicted brain lesion for that LLM condition.

## 2.6 Validation Against Human Lesions

To validate LLM-derived predicted lesions, we compared them to lesion profiles in humans with similar aphasia error profiles. Specifically, for each LLM perturbation condition, we identified the top $k$ = 5 humans whose error profiles were most similar to the LLM's error profile, measured using Euclidean distance on raw error proportions. The choice of k = 5 reflects a bias–variance tradeoff: smaller k yields unstable lesion estimates driven by idiosyncratic stroke anatomy, whereas larger k reduces anatomical specificity by averaging across heterogeneous aphasia phenotypes. This value is consistent with prior lesion-based validation approaches that average across small, behaviorally matched patient sets (Bates et al. 2003; Pustina et al. 2018; DeMarco and Turkeltaub 2018; Fridriksson et al. 2018). We then computed the mean lesion profile across these matched human participants and quantified its similarity to the LLM-predicted lesion profile using Pearson correlation.

To assess whether this correspondence exceeded what would be expected by chance, we constructed a random baseline for each condition by repeatedly sampling sets of five human participants at random (2,000 permutations per condition), computing the mean lesion profile for each random set, and measuring its correlation with the LLM-predicted lesion profile. This yielded a null distribution of correspondence values against which the observed error-matched lesion profile correspondence was compared.

Statistical significance was evaluated at both the condition level and the population level. At the condition level, permutation-based p-values were computed as the proportion of random-set correlations exceeding the observed correlation. At the population level, we tested whether correspondence between LLM-predicted lesions and error-matched human lesions was consistently greater than correspondence with randomly selected human lesions using binomial tests on the excess of significant conditions, and Wilcoxon signed-rank tests.

## 3. Results

### 3.1 Human-Based Symptom-to-Lesion Models Predict Lesion Location

Error-based symptom-to-lesion models predicted lesion damage to a subset of language-related JHU brain regions with moderate-to-high accuracy in held-out humans across both the PNT and WAB-R tasks (Figure 3). We derived these results using leave-one-out cross-validation, in which each regional model was trained to predict lesion load for a held-out participant using models fit on all remaining participants (PNT: N-1 = 213, WAB-R: N-1 = 183). Figure 3 shows the average predictive accuracy ($R^2$) for each ROI.

For the PNT-based models, prediction performance was the strongest in the left superior longitudinal fasciculus (SLF_L; $R^2 = 0.26$), posterior middle temporal gyrus (PSMG_L; $R^2 = 0.22$), left superior temporal gyrus (STG_L; $R^2 = 0.22$), and left posterior superior temporal gyrus (PSTG_L; $R^2 = 0.21$). Additional regions showed moderate predictive accuracy, including the left angular gyrus (AG_L), left posterior insula (PIns_L), and left middle occipital gyrus (MOG_L).

For the WAB-R-based models, prediction accuracy was likewise highest in the SLF_L ($R^2 = 0.21$), PSMG_L ($R^2 = 0.15$), and PSTG_L ($R^2 = 0.14$), with additional contributions from the left supramarginal gyrus (SMG_L) and left middle temporal gyrus (MTG_L).

Critically, three regions, the SLF_L, PSTG_L, and PSMG_L, ranked among the most reliably predicted regions in both the PNT and WAB-R analyses. This convergence across tasks indicates that error profiles derived from independent behavioral batteries encode lesion information for core dorsal–temporal language pathways. Overall error type distributions across the PNT and WAB-R are shown in Supplementary Figure S2, and rank-order correspondence of lesion–symptom associations across PNT and WAB-R error categories is reported in Supplementary Table S1. Prediction accuracy was generally higher for larger regions with greater lesion prevalence, consistent with the increased statistical power afforded by adequate lesion coverage.

These results validate that error profiles contain sufficient information to predict lesion location. The trained models can therefore be meaningfully applied to error profiles, including those generated by LLMs, establishing the foundation for BLUM cross-system mapping. Comprehensive lesion–symptom mapping results showing the neural correlates of each error type, along with additional validation analyses including complementary multivariate analyses, are provided in Supplementary Results S2.

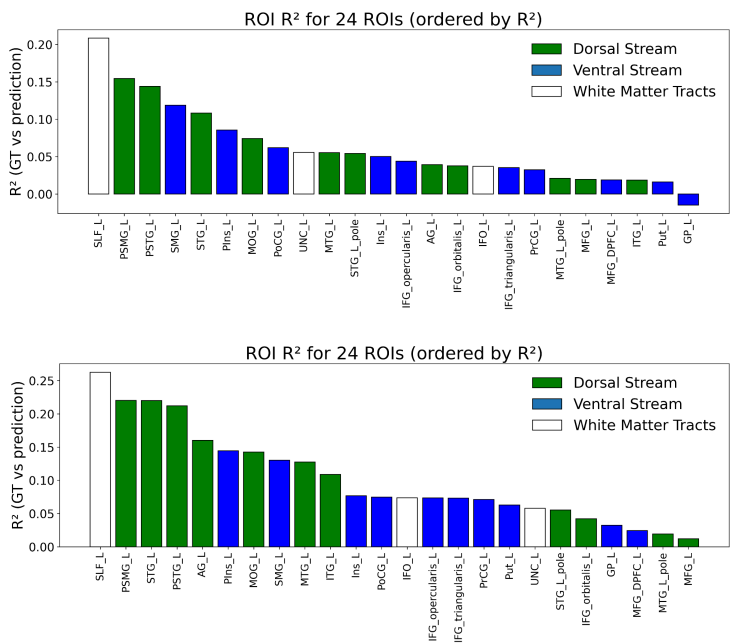

**Figure 3. Symptom-to-lesion model performance for PNT-based and WAB-R-based models.** Bar plots show $R^2$ values for each of the 24 a priori JHU atlas ROIs known to be involved in language processing. Regions are color-coded by anatomical class: dorsal stream (blue), ventral stream (green), and white matter tracts (white). Error profiles alone significantly predict lesion location with moderate-to-high accuracy, validating the foundation for mapping LLM errors into brain space. Top: PNT-based symptom-to-lesion models. Bottom: WAB-R-based symptom-to-lesion models.

## 3.2 LLM Perturbations Produce Graded Error Profiles

LLM error distributions varied systematically with perturbation severity and transformer layer (Figure 4). At low perturbation levels (10–20% modification, low noise), the model maintained near-ceiling performance with predominantly correct responses. As perturbation severity increased, correct responses declined systematically and were replaced by a characteristic progression of error types.

At moderate perturbation levels (30–50% modification), semantic and unrelated errors emerged as the dominant error types, suggesting that intermediate disruption preferentially affects lexical–semantic retrieval. At high perturbation levels (60–90% modification), no-response errors became increasingly prevalent, paralleling patterns observed in severe aphasia where retrieval failures dominate.

Perturbation effects also varied across transformer layers. Semantic errors exhibited a distinctive pattern, peaking when middle layers (approximately 8–20) were perturbed, consistent with evidence that these layers encode abstract semantic representations. Early layers (1–5) showed greater robustness to perturbation, suggesting redundancy or reduced criticality for task performance. Importantly, LLM error profiles fell within the distribution of human error profiles (Supplementary Figure S5), confirming that the two systems can be meaningfully compared. The full perturbation parameter space and its effect on error distributions are shown in Supplementary Figure S6.

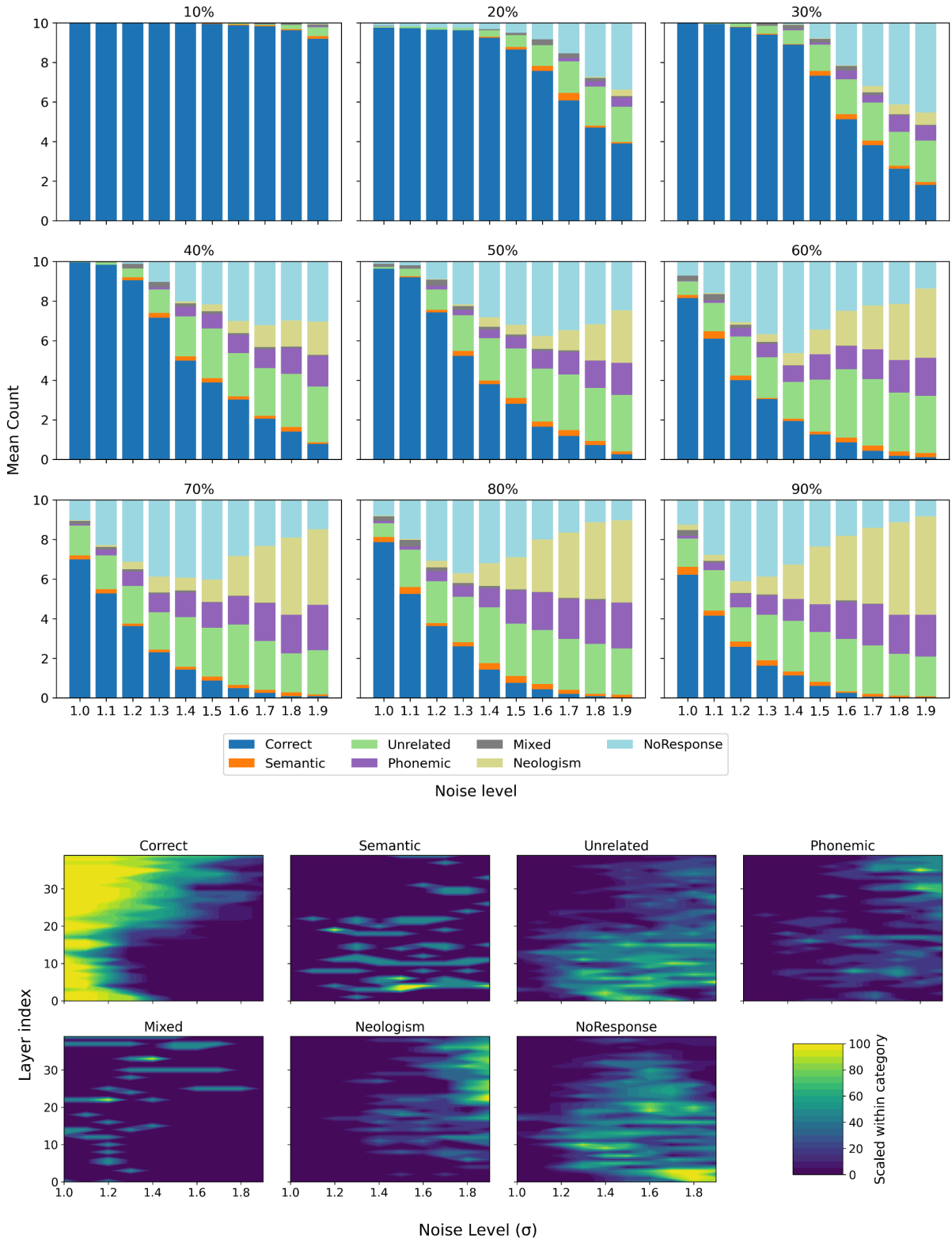

**Figure 4. LLM error distributions across perturbation conditions.** Top: Distribution of error types as a function of modification percentage (proportion of weights perturbed). Bottom: Distribution of error types as a function of the perturbed transformer layer (1–40). These graded profiles parallel patterns observed in human aphasia. For example, semantic errors peak at moderate perturbation levels and in middle layers, whereas no-response errors dominate at severe perturbation levels.

**3.3 LLM-Derived Predicted Lesions Correspond to Matched Human Lesions**

**Validation of LLM-predicted lesion profiles against human data.**

We evaluated whether lesion profiles predicted from LLM error behavior corresponded more closely to lesion profiles observed in humans with similar spoken error patterns than expected by chance. For each LLM perturbation condition (PNT: N = 1,987; WAB-R: N = 2,228), we identified the five human participants whose error profiles were most similar to the LLM's error profile (measured by Euclidean distance on raw error proportions) and computed the mean lesion profile across these five matched participants. We then compared the similarity between this error-matched human lesion profile and the LLM-predicted lesion profile to a random baseline obtained by repeatedly sampling sets of five humans uniformly at random from the full participant pool, irrespective of error similarity, and averaging their lesion profiles.

**Naming errors.**

For the PNT, correspondence between LLM-predicted lesion profiles and lesion profiles of error-matched humans exceeded chance expectations across conditions. Across all perturbations, matched correspondence exceeded the random baseline in 1,333 of 1,987 conditions (67.1%), a proportion significantly higher than expected under the null hypothesis of random matching (binomial test, one-sided p = $5.23 \times 10^{-24}$).

At the distributional level, a Wilcoxon signed-rank test confirmed that the median improvement in lesion-profile similarity relative to the random baseline exceeded zero (median $\Delta r = 0.073$; one-sided $p = 6.93 \times 10^{-26}$). Although the mean improvement in similarity was modest ($\Delta r = 0.013$), the consistency of positive differences across conditions indicates a systematic alignment between LLM error behavior and human lesion patterns for the PNT (Figure 5).

**Sentence completion errors.**

For the WAB-R, correspondence between LLM-predicted lesion profiles and error-matched human lesions was significantly stronger. Matched correspondence exceeded the random baseline in 1,522 of 2,228 conditions (68.3%), again significantly above chance (binomial test, one-sided $p = 1.25 \times 10^{-68}$).

The magnitude of this effect exceeded that observed for the PNT. A Wilcoxon signed-rank test revealed a robust positive shift in lesion-profile similarity relative to the random baseline (median $\Delta r = 0.081$; one-sided $p = 2.62 \times 10^{-92}$), with a mean improvement of $\Delta r = 0.071$. These results indicate that LLM error patterns on the WAB-R map more strongly and more consistently onto anatomically grounded lesion patterns observed in humans (Figure 5).

**Comparison across tasks.**

Together, these findings demonstrate that LLM error behavior contains anatomically meaningful structure that aligns with human lesion–symptom relationships across tasks, with substantially stronger and more consistent alignment for the WAB-R than for the PNT. This difference implies that the WAB-R elicits error patterns that more directly reflect underlying neuroanatomical constraints, whereas PNT performance may involve greater variability or redundancy across anatomical substrates.

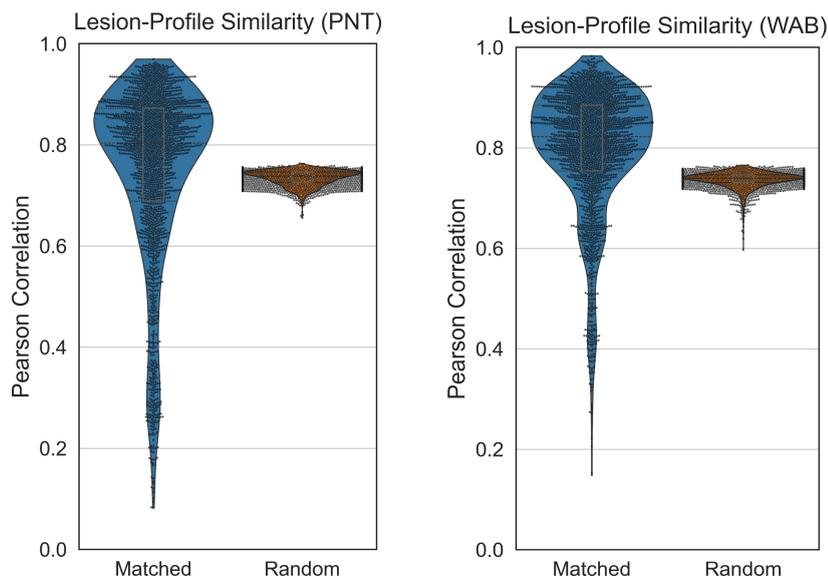

**Figure 5. Model–human lesion similarity relative to a matched random baseline.** For each LLM condition, we compared the model's predicted lesion profile to human lesion data after identifying the $k = 5$ most similar human participants in error-profile space (Euclidean distance on raw error proportions, without normalization). Violin plots show the distribution across conditions of the Pearson correlation between the LLM-predicted lesion profile and the mean lesion profile of the error-matched humans (Matched), alongside a matched random baseline (Random) computed by repeatedly sampling random sets of five humans and averaging their lesion profiles. Central lines indicate the median and interquartile range. Results are shown separately for the PNT and WAB-R datasets, demonstrating that similarity between model predictions and empirically matched human lesions consistently exceeds chance expectations. The broader dispersion for error-matched humans reflects condition-specific variability in diagnostic specificity, whereas correlations for randomly chosen humans form a narrow null distribution due to exchangeability and averaging.

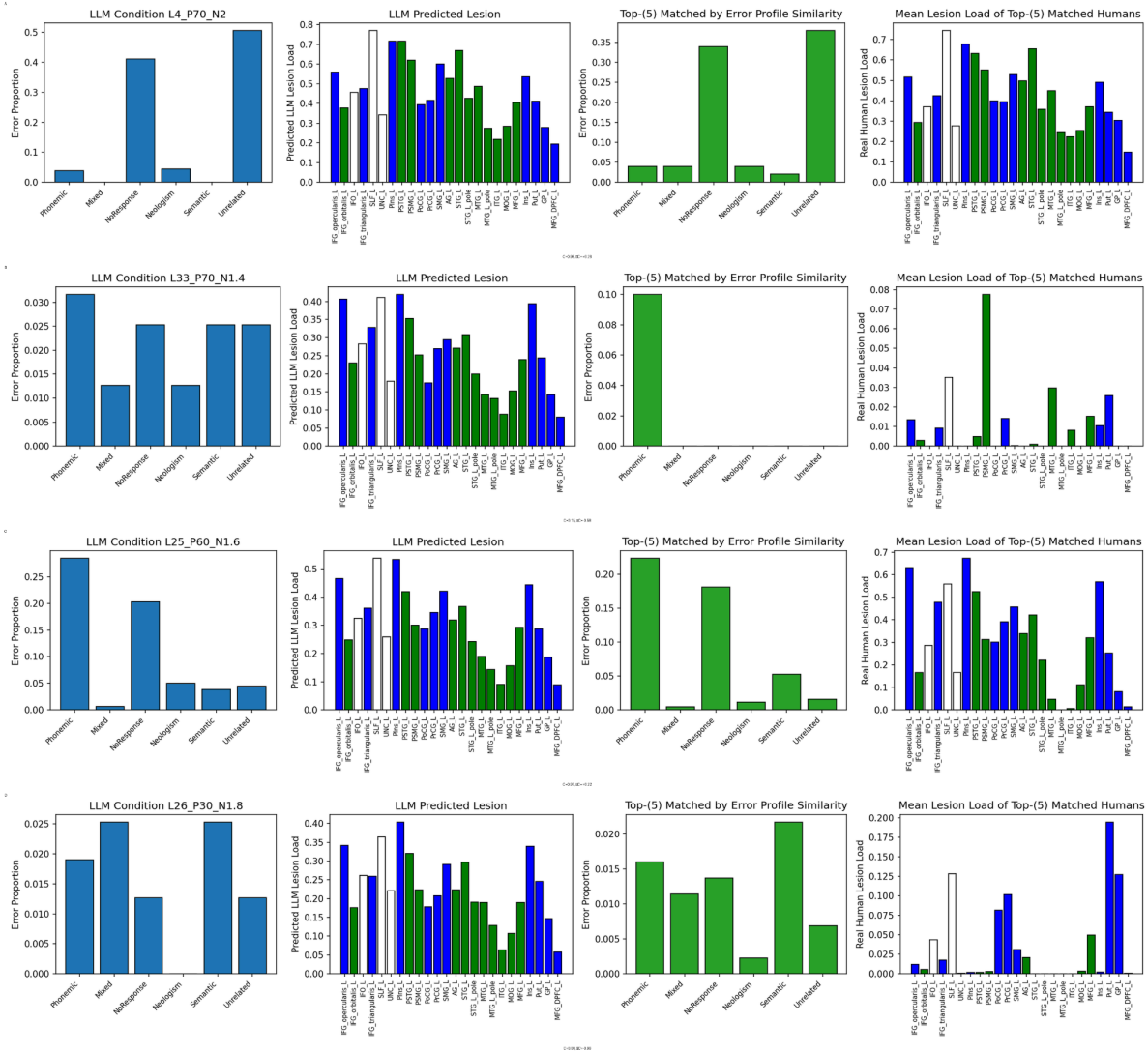

**Figure 6. Example LLM perturbation conditions illustrating low, medium, and high correspondence with human lesion profiles.** For each example condition, panels show (left to right): (1) the LLM error profile (proportions of each error type), (2) the lesion profile predicted from the LLM's errors using a symptom-to-lesion model trained on human data, (3) the average error profile of the top-k human participants whose error profiles most closely matched the LLM's error profile, and (4) the mean lesion profile of those same matched human participants. Lesion profiles are shown across 24 anatomically defined ROIs, labeled by JHU atlas region names and color-coded by anatomical class (dorsal stream = blue, ventral stream = green, white matter tracts = white). Absolute similarity between predicted and matched human lesion profiles is quantified using the Pearson correlation coefficient (C). The top two rows show example WAB-R conditions with strong vs. weak absolute similarity; the bottom row shows the same comparison for the PNT. Filenames include both absolute similarity (C) and specificity relative to chance (ΔC). Row 1: WAB-R, high similarity (L8.0_P80.0_N1.9; C = 0.97). Row 2: WAB-R, low similarity (L28.0_P80.0_N1.3; C = 0.21). Row 3: PNT, high similarity (L25.0_P60.0_N1.6_C0.97; C = 0.97). Row 4: PNT, low similarity (L33.0_P50.0_N1.6_C0.08; C = 0.08).

## 3.4 Anatomical Consistency with Dual-Stream Organization

We tested whether LLM-derived predicted lesions respected the dual-stream organization established by human lesion−symptom mapping. We classified LLM conditions with high semantic error proportions as 'semantic-dominant' and conditions with high phonemic error proportions as 'phonemic-dominant'.

For each LLM condition, we computed a ventral–dorsal stream index defined as the mean predicted lesion load across ventral ROIs minus the mean predicted lesion load across dorsal ROIs. We ranked conditions by a semantic−phonemic error score. As a sensitivity analysis, we compared the top 200 semantic-dominant and top 200 phonemic-dominant LLM conditions using a permutation test (50,000 permutations) on the difference in mean stream index; we additionally report Mann–Whitney $U$ and Welch $t$-tests.

Restricting the analysis to extreme-condition comparisons (top 200 semantic-dominant vs. top 200 phonemic-dominant), both the PNT and WAB-R exhibited significantly greater ventral bias for semantic-dominant conditions (PNT mean difference = 0.027, permutation p = $2 \times 10^{-5}$, Cohen's $d \approx 1.37$; WAB-R mean difference = 0.056, permutation p = $2 \times 10^{-5}$, Cohen's $d \approx 3.38$).

This anatomical dissociation replicates patterns established by decades of human lesion–symptom mapping, suggesting that LLM error patterns reflect processing distinctions that parallel the functional organization of human language (Indefrey and Levelt 2004).

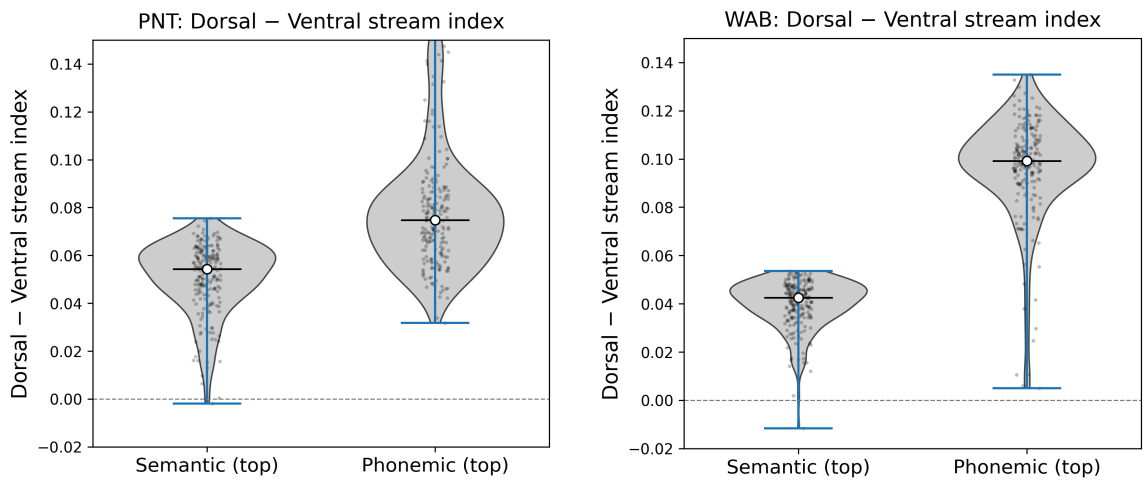

**Figure 7. Evidence for dual-stream (dorsal−ventral) organization for phonemic vs. semantic extreme conditions in LLMs.** Violin plots show the distribution of the ventral–dorsal stream index for the top 200 semantic-dominant and top 200 phonemic-dominant LLM conditions. Lower values indicate relatively greater predicted damage in dorsal-stream brain regions, whereas higher values indicate relatively greater predicted damage in ventral-stream regions. Left: PNT (mean difference = 0.027; permutation $p = 2 \times 10^{-5}$). Right: WAB-R (mean difference = 0.056; permutation $p = 2 \times 10^{-5}$). Dots indicate medians; jittered points show individual conditions. These comparisons indicate that LLM-predicted lesion profiles exhibit ventral-stream bias for strongly semantic conditions and

dorsal-stream bias for strongly phonemic conditions, with a larger effect size observed for WAB-R.

## 3.5 Divergences Between Biological and Artificial Systems

Several divergences emerged between biological and artificial systems. Phonemic errors occurred substantially less frequently in LLM perturbations than in human aphasia, suggesting that transformer architectures may implement more robust phonological encoding or that their failure modes preferentially affect semantic rather than phonological processing. One likely source of this divergence is a difference in representational units and processing stages. Human language production involves explicit phonological encoding and articulatory sequencing, supported by dorsal-stream mechanisms, such that damage frequently yields phonemic paraphasias and neologisms. In contrast, transformer-based LLMs operate over discrete lexical or sublexical tokens that are optimized for statistical prediction rather than phonological assembly. As a result, perturbations in LLMs are more likely to manifest as semantic or lexical substitution errors than as phonological assembly errors, which have no direct computational analogue in current token-based architectures. No-response errors emerged at extreme perturbation levels, but their layer-wise gradients differed from those predicted by human lesion patterns. The divergent layer-wise gradients observed for No Response errors warrant special consideration. In human aphasia, no-response behavior often reflects breakdowns in lexical retrieval or phonological encoding that occur despite preserved engagement with the task, and is typically associated with damage to specific dorsal or ventral stream components depending on the underlying deficit. In contrast, in LLMs, no-response errors frequently emerge at extreme perturbation levels and may reflect global disruption of generation dynamics, such as collapse of token probability distributions or failure to initiate decoding, rather than selective impairment of a linguistic subroutine. The fact that superficially similar error categories arise from different internal failure modes underscores an important strength of the BLUM framework: it does not presuppose equivalence between biological and artificial systems, but instead enables identification of where behavioral similarity masks mechanistic divergence.

These divergences highlight key differences between transformer architectures and biological language systems. Importantly, such differences are themselves informative, as they identify aspects of language processing where artificial and biological systems may rely on fundamentally different computational strategies.

## 4. Discussion

**4.1 Summary of Principal Findings**

The present study establishes a proof of concept for BLUM as a potential framework for using human clinical data as external ground truth for LLM interpretability. Three principal findings support this framework. First, symptom-to-lesion models trained on spoken error profiles from individuals with chronic post-stroke aphasia predicted lesion location with moderate-to-high accuracy, establishing that behavioral error patterns encode sufficient neuroanatomical information to reconstruct cortical damage location (Bates et al. 2003; Pustina et al. 2017; DeMarco and Turkeltaub 2018). Second, systematic perturbations to transformer layers produced graded error profiles that fell within the distribution of human error profiles, confirming that the two systems can be meaningfully compared using a common error classification taxonomy (Patterson et al. 1994; Plaut 1996). Third, lesion profiles predicted from LLM error patterns corresponded to actual lesions in humans with similar error profiles above chance in 67% of picture-naming conditions and 68.3% of sentence-completion conditions; semantic-dominant LLM errors predicted ventral-stream lesions, and phonemic-dominant errors predicted dorsal-stream lesions (Hickok and Poeppel 2007; Fridriksson et al. 2018; Mirman et al. 2015). Together, these findings demonstrate that human lesion–symptom mapping can serve as external ground truth for understanding LLM organization.

**4.2 Human Lesion Data as External Ground Truth for LLM Interpretability**

BLUM addresses a fundamental challenge in LLM interpretability: the absence of external validation criteria. Current approaches evaluate interpretability claims using metrics internal to the model or benchmark performance, without an independent criterion for assessing validity (Belinkov 2022; Belinkov and Glass 2019; Hewitt and Manning 2019; Conneau et al. 2018). Probing classifiers may reveal that linguistic information is recoverable from specific activations, but such recoverability does not establish that the information causally contributes to model outputs (Ravichander et al. 2021). Sparse autoencoders may identify interpretable features, but without external validation it remains unclear whether these features represent the correct decomposition (Cunningham et al. 2023; Bricken et al. 2023; Templeton et al. 2024).

BLUM leverages what is arguably the most rigorous causal framework available for understanding language function: more than a century of human lesion–symptom mapping (Broca 1861; Wernicke 1874; Geschwind 1970). The accumulated knowledge from clinical neuroscience provides detailed maps of which brain regions are necessary for specific language functions (Dronkers et al. 2004; Fridriksson et al. 2018; Mirman et al. 2015), offering precisely the kind of external reference that LLM interpretability currently lacks. The correspondence we observe between LLM-predicted lesions and actual human lesions demonstrates that this form of external validation is achievable. This alignment is not guaranteed by the method; rather, it constitutes an empirical finding that validates the use of human clinical data as an external reference for LLM interpretability

## 4.3 Task Differences in Cross-System Alignment

The substantially stronger correspondences observed for sentence completion compared to picture naming likely reflect differences in task constraints and the mechanistic sources of errors. In humans, sentence completion inherently narrows the space of plausible responses for each item, reducing variability in error patterns and facilitating more consistent mapping between LLM perturbation effects and human lesion profiles. In contrast, picture naming can yield errors at multiple processing stages (Levelt et al. 1999; Dell et al. 1997). For example, failure to produce the correct word could reflect impaired visual recognition, degraded semantic access, failed lexical retrieval, or disrupted phonological encoding, each of which may recruit distinct neural substrates (Indefrey and Levelt 2004; Hickok and Poeppel 2007). This mechanistic complexity introduces variability that may not map cleanly onto LLM perturbation effects, suggesting that task selection should be carefully considered when extending BLUM to new domains.

## 4.4 Behavioral Consistency with Dual-Stream Organization

The observed dissociation between semantic-dominant LLM conditions, which predicted ventral-stream lesions, and phonemic-dominant conditions aligns with the dual-stream organization of human language processing (Hickok and Poeppel 2007; Rauschecker and Scott 2009; Saur et al. 2008). However, this finding should be interpreted with appropriate caution. The symptom-to-lesion models were trained on human patients whose error patterns already reflect dual-stream organization (Fridriksson et al. 2016), raising the possibility that the anatomical dissociation in LLM-predicted lesions is inherited from the training data rather than reflecting intrinsic organizational properties of the LLM itself. Nevertheless, the correspondence demonstrates that LLM error patterns contain sufficient structure to be meaningfully projected into neuroanatomically organized space. Future work can strengthen these findings by integrating BLUM with direct analysis of representational geometry within transformer layers (Geiger et al. 2021; Wu et al. 2023), enabling tests of whether behavioral alignment with human lesion patterns corresponds to intrinsic organizational properties within LLM architectures.

## 4.5 Divergences Between Biological and Artificial Systems

The relative scarcity of phonemic errors in perturbed LLMs compared to human aphasia likely reflects fundamental differences in representational units. In humans, phonemes or syllables constitute the basic units of language processing, and the dorsal stream supports the assembly of these units into articulable sequences (Hickok and Poeppel 2007; Hickok 2012). LLM tokenizers, by contrast, represent language primarily as whole words or subword units such as morphemes and word stems (Sennrich et al. 2016; Kudo and Richardson 2018). When LLM layers are perturbed, failures are therefore more likely to manifest as semantic or lexical errors rather than as phonological errors involving incorrect assembly of sound sequences. This pattern contrasts with human aphasia, in which phonemic paraphasias and neologisms are common consequences of dorsal-stream damage (Buchsbaum et al. 2011; Fridriksson et al. 2018). Rather than representing a limitation of the BLUM framework, this divergence is scientifically informative, as it highlights an aspect of language processing where biological and artificial systems employ fundamentally different representational strategies, delineating the boundary conditions of cross-system comparison (Chen et al. 2024; Faries and Raja 2022).

## 4.6 Implications and Future Directions

BLUM provides a principled external criterion for evaluating interpretability claims. If activation analyses suggest that a particular set of middle transformer layers encodes semantic information (Meng et al. 2022; Zhang and Nanda 2023), BLUM can test this claim by examining whether targeted perturbation of these layers produces error profiles enriched for semantic errors and whether the resulting predicted lesions concentrate in ventral-stream regions associated with semantic processing in humans (Binder et al. 2009; Mirman et al. 2015). If perturbation instead produces diffuse error patterns that fail to map onto coherent lesion predictions, the original interpretability claim warrants skepticism. Conversely, if perturbation yields anatomically specific predictions that align with established human lesion–symptom relationships, BLUM provides external validation that complements internal activation analyses (Geiger et al. 2021; Wu et al. 2023). Most critically, BLUM as presented validates behavioral alignment but leaves open whether this reflects convergent internal organization. A direct test would reverse the mapping: taking human patient error profiles and identifying which LLM perturbation conditions produce maximally similar errors. If patients with semantic deficits consistently map onto middle-layer perturbations while patients with phonological deficits map onto distinct layers, this would constitute evidence that LLM organization recapitulates functional distinctions that are neuroanatomically grounded in humans, transforming BLUM from a behavioral validation framework into a bidirectional mapping between biological and artificial systems.

BLUM also offers a principled approach to model compression grounded in functional necessity rather than parameter magnitude (Han et al. 2015; Ma et al. 2024). Current compression methods rely on magnitude-based or gradient-based heuristics and lack principled criteria for distinguishing truly dispensable parameters from those supporting capabilities not captured by existing benchmarks (Sun et al. 2023; Kim et al. 2024). If perturbation of early transformer layers produces minimal deficits across a range of severity levels, with predicted lesion profiles that fail to correspond to human lesions, these layers may represent candidates for pruning or aggressive quantization (Men et al. 2025; Gromov et al. 2025). In contrast, if perturbation of middle layers produces severe deficits even at low severity and yields predicted lesions that strongly correspond to human data, these layers should be preserved. By systematically characterizing perturbation effects across layers and mapping the resulting error profiles into brain space, BLUM enables construction of a functional importance map that distinguishes critical hubs from redundant components (Chen et al. 2024; Song et al. 2024). Future work should develop compression algorithms that leverage these BLUM-identified dispensable components, systematically removing parameters whose perturbation produces minimal functional deficits while preserving circuits whose perturbation yields severe and specific impairments.

In clinical neuroscience, BLUM opens the possibility of creating patient-specific digital twins for conditions such as aphasia and dementia (Broderick et al. 2022; Laubenbacher et al. 2021). By identifying LLM perturbation conditions that replicate individual patient error profiles, researchers can generate computational surrogates that simulate patient-specific deficits. Archival databases contain rich clinical data beyond error profiles—lesion maps, treatment histories, comorbidities, and longitudinal assessments—that could serve as conditioning variables for multivariate patient-to-model matching. For progressive conditions like dementia, systematic increases in perturbation severity may simulate disease trajectory, generating testable predictions about functional decline. Digital twins would enable experiments impractical in human participants: simulating thousands of virtual patients to identify treatment moderators, testing intervention parameters in silico, and generating synthetic datasets for machine learning applications (Plaut 1996; Welbourne and Lambon Ralph 2007). Validating such twins against held-out patient outcomes would establish clinical utility and potentially transform how intervention research is conducted.

Finally, extending BLUM to multiple LLM architectures is necessary to test generalization and identify architecture-specific organizational principles. Applying BLUM to GPT, LLaMA, Claude, and related model families would reveal whether correspondence with human lesion patterns represents a general property of transformer-based language models or varies across architectures (Tuckute et al. 2024; Oota et al. 2023). The aphasia database used here also contains extensive behavioral data beyond the tasks examined in this study, including assessments of auditory comprehension (Kertesz and Raven 2007), syntax (Caplan et al. 2007), reading (Coltheart et al. 2001), and writing (Rapcsak and Beeson 2004), as well as measures of non-verbal cognitive abilities (Helm-Estabrooks 2001; Murray 2012). These data enable testing whether BLUM generalizes beyond spoken language production, potentially revealing whether the framework captures broader aspects of cognition.

### 4.7 Limitations

Several limitations warrant consideration across the three stages of the BLUM pipeline. At the foundation stage, symptom-to-lesion models were trained on a single database of English-speaking participants from one research center; generalization to other languages, etiologies, and populations requires further testing (Swinburn et al. 2004; Ivanova et al. 2021). At the translation stage, we evaluated a single model architecture, LLaVA-1.6-Vicuna-13B (Liu et al. 2023), and extension to other architectures remains an empirical question. The assumption that error profiles are comparable across biological and artificial systems despite fundamental architectural differences is strong (Tuckute et al. 2024; Antonello et al. 2024), and tokenization strategies employed by LLMs may limit certain error types (Sennrich et al. 2016). At the validation stage, we employed a top-k matching approach with k = 5 and Pearson correlation as the similarity metric; alternative matching strategies or similarity measures might yield different results (Kriegeskorte et al. 2008). In addition, validation was limited to two tasks, capturing only a subset of language functions. Despite these limitations, the consistent above-chance correspondence observed across nearly 2,000 perturbation conditions, combined with anatomically appropriate dual-stream dissociations, establishes proof of concept for the BLUM framework.

## 5. Conclusions

This study demonstrates that human lesion–symptom mapping—the gold standard for establishing causal brain-behavior relationships for over a century—can serve as an external reference structure for evaluating LLM perturbation effects. By applying identical clinical assessments to stroke patients and perturbed LLMs and classifying responses using a common error taxonomy, we showed that LLM error profiles can be meaningfully projected into human neuroanatomical space.

Three findings support this framework: symptom-to-lesion models predicted lesion location from error profiles alone; LLM perturbations produced error distributions comparable to human aphasia; and LLM-predicted lesions corresponded to actual lesions in error-matched humans, with semantic errors mapping onto ventral-stream lesion patterns and phonemic errors onto dorsal-stream patterns consistent with the organization learned from human patients. These findings validate behavioral alignment between perturbed LLMs and human aphasia, while leaving open whether such alignment reflects convergent internal organization—a question that motivates direct investigation through reverse mapping of human error profiles onto LLM perturbation space. For AI development, BLUM provides a principled approach to model compression based on functional necessity and offers external validation for interpretability claims. For clinical neuroscience, the framework opens the possibility of creating patient-specific digital twins that incorporate rich clinical data to simulate individual deficits, model disease progression, and predict treatment responses without requiring participation in costly trials.

## Declarations

### Competing Interests

### Funding

This research was supported by a SmartState endowment established with private funding arranged by the state of South Carolina for J.F. The funder had no role in study design, data collection, and analysis, decision to publish, or preparation of the manuscript.

### Data Availability

Behavioral and demographic data will be made available upon reasonable request to the corresponding author. Lesion data cannot be shared publicly due to patient privacy constraints, but are available to qualified researchers through data use agreements.

# Supplementary Materials

## Supplementary Methods

### S1. Error Taxonomy and Classification

We developed a unified error classification system to categorize responses from both the Western Aphasia Battery–Revised (WAB-R) and the Philadelphia Naming Test (PNT) using a common taxonomy, enabling direct comparison of error distributions across tasks. The classification scheme was adapted from standard naming-error taxonomies used in aphasia research (Schwartz et al. 2006; Dell et al. 1997) and implemented as a rule-based decision tree.

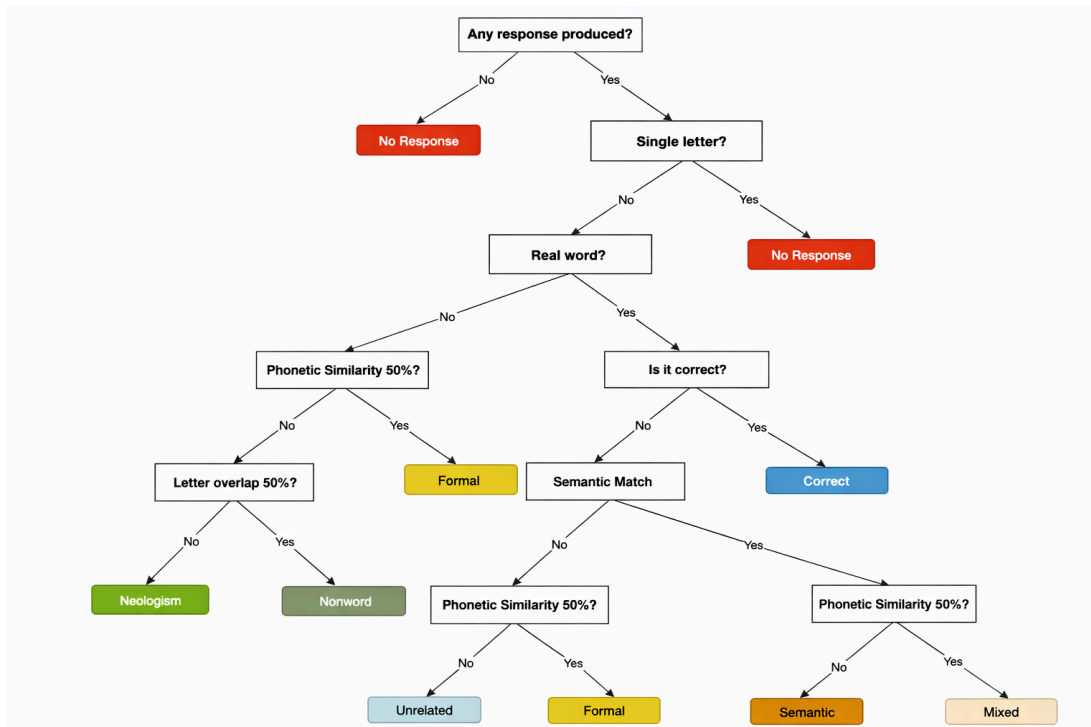

**Supplementary Methods Figure 1. Decision-tree for classification of naming responses.** This flowchart illustrates the step-by-step procedure used to classify naming responses based on response presence, lexical status, phonetic similarity, orthographic overlap, and semantic relatedness.

Responses were first categorized based on whether the participant produced a scorable verbal response. Failures to respond within the allotted time or explicit statements of inability (e.g., "I don't know") were classified as No Response errors. All other responses were categorized as Produced Word and subjected to further classification.

For produced responses, the decision tree first determined whether the output was a real English word. Real-word responses were then evaluated for correctness: responses matching the target or an accepted synonym were classified as Correct. Incorrect real-word responses were assessed for semantic relatedness to the target using two normative databases: the Florida Associative Norms (Nelson et al. 2004) and a taxonomic similarity dataset. Responses with a documented semantic relationship to the target (associative, categorical, or functional) were classified as Semantic errors (e.g., "goat" for sheep). Real-word responses that were both semantically related and phonologically similar to the target ($\geq$50% phoneme overlap) were classified as Mixed errors (e.g., "rat" for cat). Real-word responses with no identifiable semantic relationship to the target were classified as Unrelated errors if phonologically dissimilar, or as Formal errors if they shared $\geq$50% phonological similarity with the target (e.g., "sheep" for sheet).

Nonword responses were classified based on their phonological and orthographic similarity to the target. Nonword errors exhibited $\geq$50% phoneme or letter overlap with the target (e.g., "sheeb" for sheep), representing phonological distortions that retained substantial similarity to the intended word. Neologisms exhibited <50% overlap with the target, representing severe phonological disruptions that obscured the relationship to the intended response. Following standard practice in the aphasia literature, Formal and Nonword errors were consolidated into a single Phonemic error category for analyses requiring comparison with prior lesion-symptom mapping studies.

All responses were independently coded by two trained raters: a speech-language pathologist with clinical experience in aphasia assessment and a linguist specializing in lexical-semantic processing. Each rater independently categorized every response according to the decision tree. Inter-rater discrepancies were resolved through discussion, and consensus labels were used for all subsequent analyses.

To facilitate reliable classification of large datasets and ensure reproducibility, we developed an automated scoring system implementing identical decision-tree criteria. Phonological similarity was computed algorithmically using phoneme-level Levenshtein distance, with phonemic transcriptions derived from the Carnegie Mellon University Pronouncing Dictionary. Semantic relatedness was determined by querying the Florida Associative Norms and taxonomic similarity databases. The automated system achieved >97% agreement with consensus human annotations. Disagreements occurred most frequently for responses near classification boundaries, particularly between Semantic and Unrelated categories (27 discrepancies in 908 error responses).

## S2. Neuroimaging Acquisition and Lesion Delineation

Structural magnetic resonance imaging (MRI) data were acquired for all participants using T1-weighted and T2-weighted sequences. Lesion masks were generated following protocols established in prior large-scale lesion–symptom mapping studies of aphasia (Fridriksson et al., 2018). Lesions were manually demarcated on T2-weighted images by a trained neurologist, as T2 sequences provide superior contrast for identifying chronic stroke damage. Manual delineation remains the gold standard for lesion identification, ensuring accurate capture of infarct boundaries that automated methods may miss or misclassify.

Lesion masks were normalized to Montreal Neurological Institute stereotactic space using enantiomorphic normalization (Nachev et al. 2008), a procedure that addresses distortion artifacts that can arise when normalizing brains with large lesions. This approach replaces damaged tissue with mirrored tissue from the intact contralateral hemisphere before spatial normalization, yielding more accurate alignment with the template brain. The normalization pipeline was implemented using our open-source nii_preprocess toolbox.

Normalized lesion maps were intersected with the Johns Hopkins University (JHU) atlas. This widely used parcellation scheme provides anatomically defined regions of interest (ROIs) spanning cortical gray matter and white matter tracts. The JHU atlas includes 189 labeled structures, enabling systematic quantification of lesion burden across the brain.

The present analysis focused on 24 left-hemisphere ROIs selected based on prior work implicating these regions in language processing (Fridriksson et al., 2018). These included cortical regions associated with the dorsal stream (inferior frontal gyrus pars opercularis and triangularis, precentral gyrus, postcentral gyrus, supramarginal gyrus, superior temporal gyrus) and ventral stream (middle temporal gyrus, inferior temporal gyrus, fusiform gyrus, angular gyrus), as well as the insula and adjacent structures. We expanded the anatomical space beyond prior work to include three white-matter tracts strongly implicated in language function: the superior longitudinal fasciculus, uncinate fasciculus, and inferior fronto-occipital fasciculus.

## S3. Lesion–Symptom Mapping Statistical Approach

Lesion–symptom mapping analyses were conducted to identify which brain regions are associated with each error type. We employed two complementary statistical approaches: mass-univariate regression and multivariate stepwise regression.

**Mass-univariate regression.**

For each behavioral measure (error counts per category, converted to z-scores), we fit separate ordinary least squares regression models predicting behavioral performance as a function of lesion load in each ROI. This approach tests the association between damage to each region and each behavioral outcome while remaining agnostic to relationships among regions. Statistical significance was assessed using permutation testing, a nonparametric approach that controls for the spatial autocorrelation and non-normal distributions common in lesion data. For each behavioral measure, 400 permutations were conducted, randomly shuffling the mapping between lesion data and behavioral scores to generate an empirical null distribution. We employed one-tailed tests in the negative direction, consistent with the hypothesis that greater lesion burden predicts poorer performance. Resulting p-values were corrected for multiple comparisons across ROIs using the Benjamini–Hochberg false discovery rate (FDR) procedure.

**Multivariate stepwise regression.**

To identify the subset of regions jointly predicting each behavioral measure while accounting for shared variance among correlated predictors, we also employed forward–backward stepwise regression. This procedure iteratively adds and removes predictors based on predefined probability thresholds (p_enter = 0.05; p_remove = 0.10). A region enters the model only when it significantly reduces residual variance; a region is removed if its contribution becomes nonsignificant after other predictors are incorporated. This multivariate approach complements mass-univariate lesion–symptom mapping by addressing collinearity among neighboring regions that share vascular supply and are often damaged together.

## S4. Simulated Large Language Model "Lesions" Through Systematic Perturbation

In an effort to better understand similarities between human brains and large language models (LLMs), we attempt to disrupt a fully trained LLM in a way that mirrors lesions seen in an aphasic brain. We apply noise to an LLM in a systematic way and measure the impact of these artificial lesions by characterizing incorrect responses according to error categories created by speech-language pathologists.

We use the 13-billion-parameter Vicuna model, which is a transformer-based model (Chiang et al. 2023). Like the generative pre-trained transformer (GPT) models, it is autoregressive, meaning the model makes a prediction of the next token given the previous token (Radford et al. 2018). We use this model to simulate a speech-language pathologist administering the WAB-R test to an artificial patient. We use a prompt that is based on the original WAB-R prompt but includes extra suggestions such as not responding with more than one word. We then give the sentence or question to the model and record the first word that the model responds with, allowing for typical model behaviors such as repeating the prompt. Supplementary Methods Figure 2 below shows the method used to collect responses from the model.

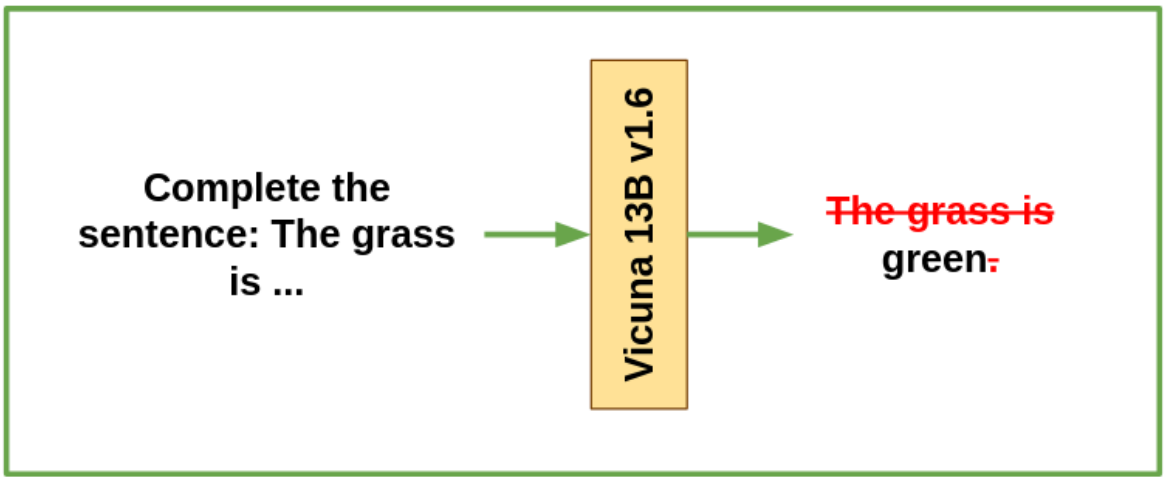

**Supplementary Methods Figure 2. Vicuna model prompting procedure.** Example illustrating how the model is prompted with a question from the WAB-R dataset and how the first word is selected after parsing out model artifacts.

The Vicuna model is a fine-tuned LLaMA model (Touvron et al. 2023). The LLaMA architecture used in the Vicuna model is shown in Supplementary Methods Figure 3 below. It has 40 multihead attention layers, each with 40 attention heads and a hidden size of 5,120, meaning each head dimension is size 128. We apply the artificial lesion to the weights in these 40 multihead attention layers. Noise may be applied to any of the weights in the LLaMA decoder layers.

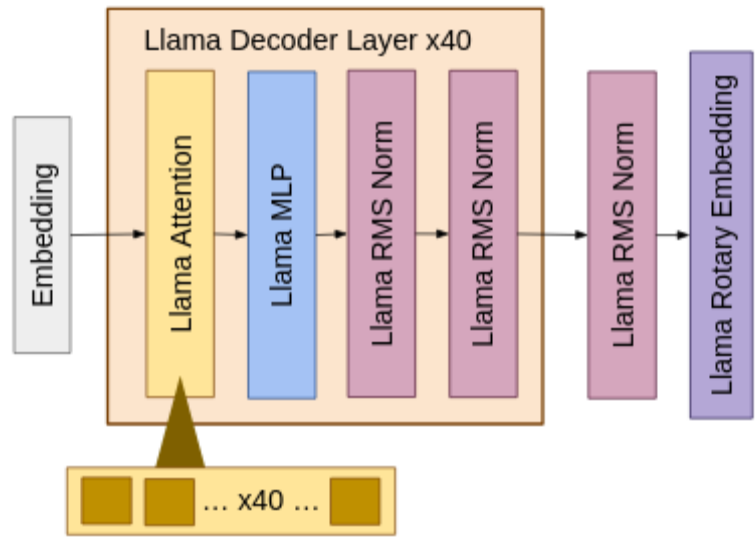

**Supplementary Methods Figure 3. LLaMA model architecture.** The model consists of 40 LLaMA decoder layers, each of which includes an LLaMA attention module, which is a multihead attention layer with 40 attention heads.

To add the artificial lesion, we use three parameters to control the amount and location of noise added. We control the level of noise added, the percentage of nodes in a layer that are modified, and the layer to which this noise is applied. We apply the percentage of nodes modified from 10–90% in 10% increments. For each modification percentage, we apply a noise standard deviation in the range of 0–1.9 with a step size of 0.1. Finally, these changes are applied to each of the 40 target layers separately. This process creates 7,200 different models with artificial lesions of different sizes in different locations.

Responses from the model on the WAB-R dataset were classified into to seven categories, including "Correct" and the error types: "Unrelated," "Semantic," "Formal," "Mixed," "Neologism," and "No Response." These classifications match the response classifications of actual patients on the WAB-R dataset, allowing direct comparison.

Once classified, the artificial patients (LLM output) and actual patients can be matched based on their responses over the 10 items in the WAB-R dataset. For example, the model instance with no artificial lesioning and the healthy patient are a match, with both "patients" scoring "correct" on every item. Out of 410 aphasic patients, 78 M patients and 10 SPARC patients had a response profile that exactly matched that of an artificial patient. Note that the quantity of responses of a certain type is considered, but the response type is not matched per individual item. Supplementary Methods Figure 4 below shows selected matching artificial and actual patients.

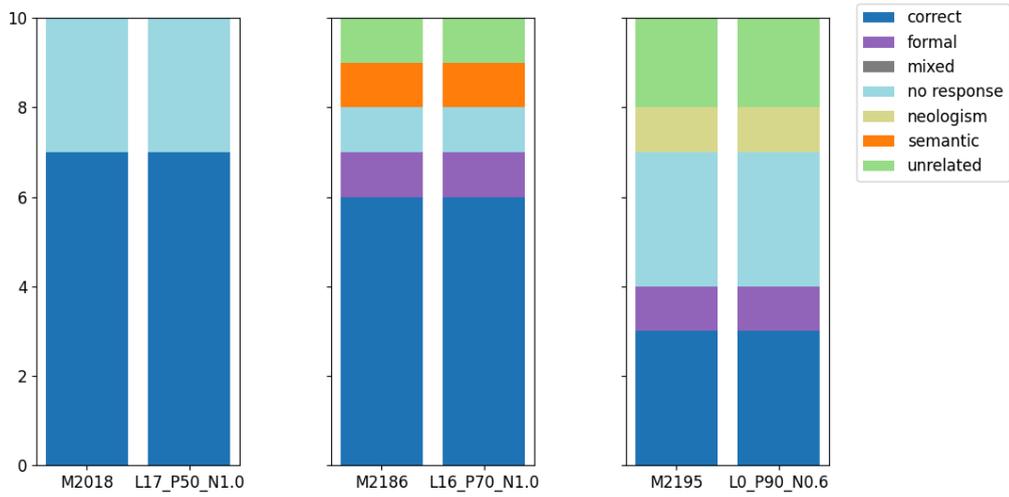

**Supplementary Methods Figure 4. Selected response profiles of matching artificial and human patients.** Across the WAB-R dataset, artificial and human patients produce responses classified into the same response categories.

## Supplementary Tables

## Supplementary Table S1. Cross-task correspondence in lesion–symptom patterns.

For each error category, 24 left-hemisphere ROIs were rank-ordered by the strength of their lesion–behavior association, and these rankings were correlated across tasks (PNT vs. WAB-R).

| Cross-Task Pair Summary (WAB × PNT) | | | |
|---|---|---|---|
| **Error Type** | **Spearman ρ** | **p-value** | **Interpretation** |
| **Correct** | 0.7426 | 0.000034 | strong |
| **Formal** | 0.7243 | 0.000063 | strong |
| **NR** | 0.4617 | 0.02312 | moderate |
| **Mixed** | 0.36 | 0.08399 | weak |
| **Neologism** | 0.2078 | 0.3298 | weak |
| **Semantic** | 0.6539 | 0.0005287 | strong |
| **Unrelated** | 0.1139 | 0.5961 | very weak |

Correct, Formal, and Semantic errors showed strong correspondence ($ρ > 0.65$, $p < 0.001$); No Response (NR) showed moderate correspondence ($ρ = 0.46$, $p < 0.05$); Mixed, Neologism, and Unrelated errors showed weak correspondence ($ρ < 0.37$, $p > 0.05$).

## Supplementary Results

### S1. Extended Symptom-to-Lesion Model Validation

We compared linear regression (LR) and support vector regression (SVR) models for symptom-to-lesion prediction. Both approaches yielded comparable prediction accuracy, with linear regression performing slightly better for most regions. This suggests that the relationship between behavioral profiles and lesion location is predominantly linear and does not require more complex nonlinear modeling, consistent with the relatively direct mapping between localized damage and specific behavioral impairments established by decades of lesion–symptom mapping research.

The spatial patterns recovered by symptom-to-lesion models converged with established relationships between aphasia phenotypes and neuroanatomy. WAB-R profiles dominated by phonological production deficits, repetition impairments, or reduced fluency elicited predicted lesion patterns concentrated in dorsal-stream regions, including premotor cortex, pars opercularis, and supramarginal gyrus. Profiles characterized by impaired comprehension, reduced information content, and difficulty with semantic retrieval produced predictions concentrated in ventral-stream regions, including the posterior and middle temporal gyri.

### S2. Full Lesion-Symptom Mapping Results

Supplementary Figure S3 presents the ROI-level lesion–symptom associations across all error types for both the PNT and WAB-R, while Supplementary Figure S4 displays the corresponding brain renderings.

Consistent with the dual-stream model of language processing (Hickok & Poeppel, 2007; Fridriksson et al., 2016), lesion–symptom associations revealed a clear anatomical dissociation between error types. Error categories associated with disrupted phonological encoding (Formal errors and neologisms) showed robust associations with dorsal-stream structures, including the inferior frontal gyrus (pars opercularis and pars triangularis), precentral and postcentral gyri, supramarginal gyrus, and the superior longitudinal fasciculus. The involvement of the insula and pars opercularis for Formal errors is consistent with lesion patterns that characterize conduction aphasia, in which phonological sequencing and articulatory buffering may be compromised.

Error types reflecting degraded lexical–semantic access (Semantic and Mixed errors) engaged a well-defined ventral-stream network, including the middle and inferior temporal gyri, fusiform cortex, angular gyrus, and the uncinate and inferior fronto-occipital fasciculi. Notably, PNT Semantic and WAB-R Semantic errors demonstrated nearly identical ventral-stream involvement, indicating that shared cortical substrates across tasks support semantic retrieval processes despite differences in task demands.

Unrelated responses showed more heterogeneous anatomical signatures spanning both dorsal and ventral territories, likely reflecting simultaneous degradation of semantic constraints and phonological well-formedness. Correct responses showed the inverse pattern: better performance was associated with preserved tissue in regions linked to error production.

The multivariate analyses revealed that considerably fewer regions emerged as independent predictors compared to the univariate analyses. The regions most frequently emerging as independent predictors across error types were posterior superior temporal gyrus, precentral gyrus, and posterior insula. These regions could be considered especially important hubs in the cortical speech and language network.

## Supplementary Figures

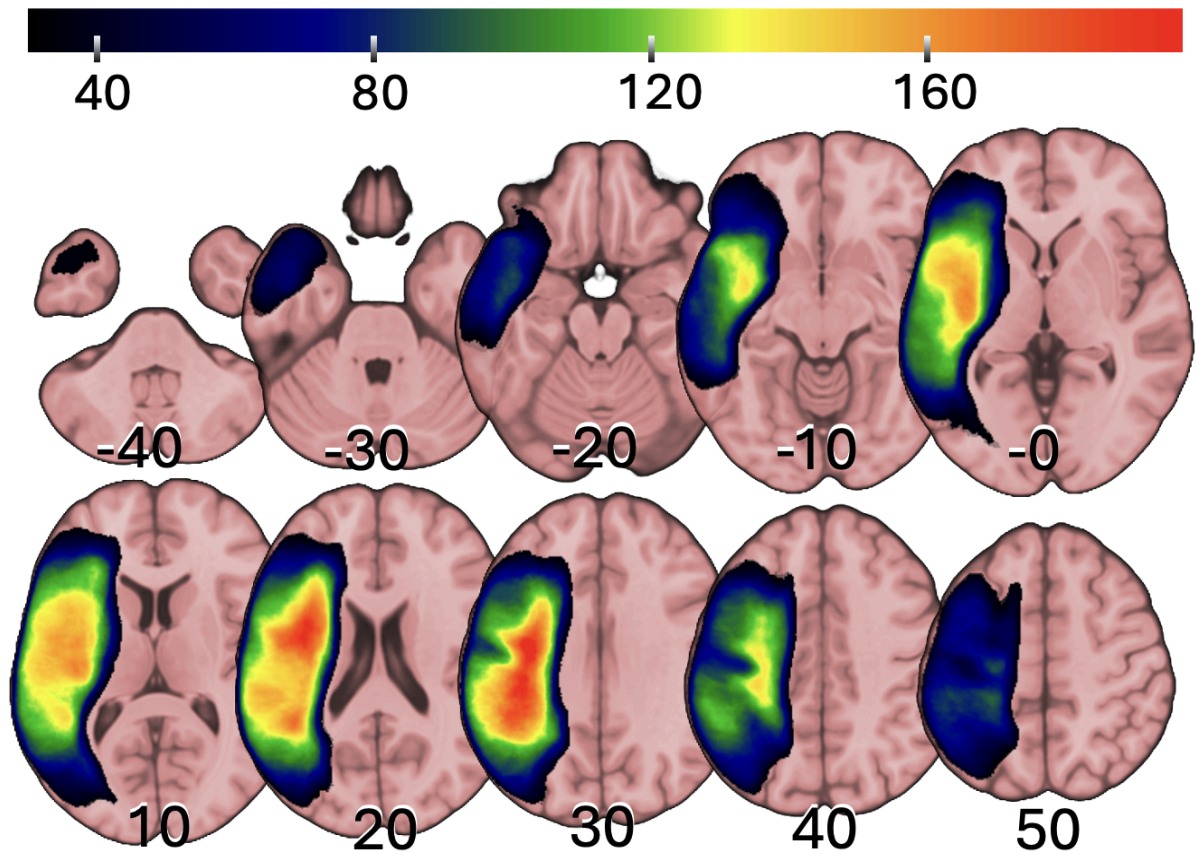

**Supplementary Figure S1. Lesion overlay map.** Voxelwise lesion overlap map for 301 participants with neuroimaging-based lesion data, overlaid on the SPM152 standard template. Lesion maps were binary at the individual level (lesion = 1, non-lesion = 0), and color-bar values in the group map represent the count of participants exhibiting a lesion at that location. Results are displayed as mosaic coronal slices. Only voxels in which at least 10% of individuals had a lesion are shown.

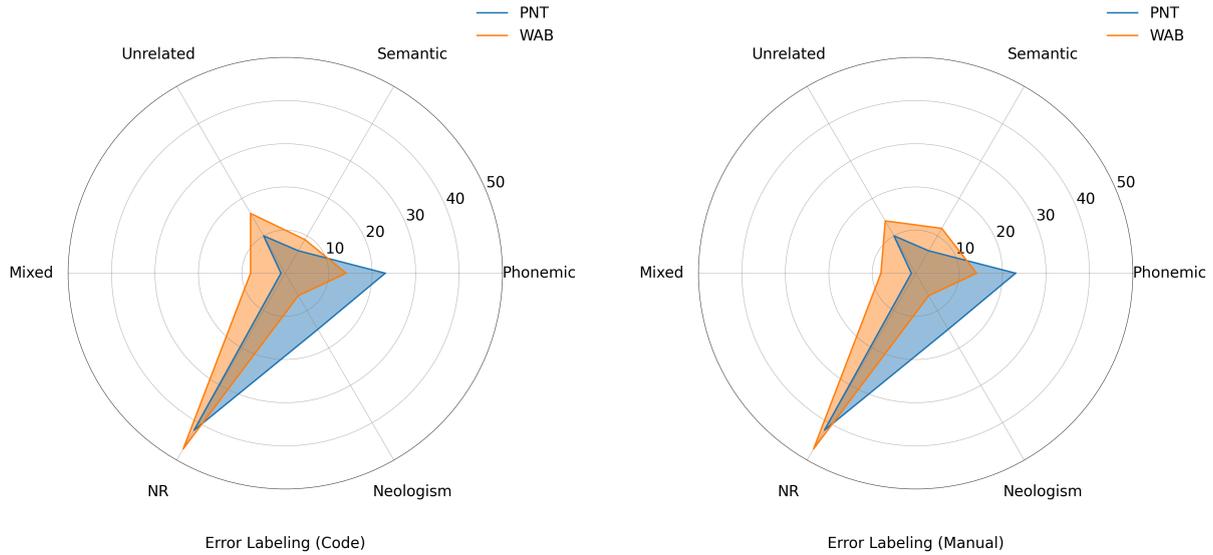

**Supplementary Figure S2. Error type distributions.** Radar plots show the percentage of responses classified into each error category for automated scoring (left) and manual scoring by clinicians (right), with PNT scores shown in blue and WAB-R scores shown in yellow. No Response errors dominate both tasks. The PNT shows higher proportions of Phonemic and Neologism errors, whereas the WAB-R shows increased Unrelated responses.

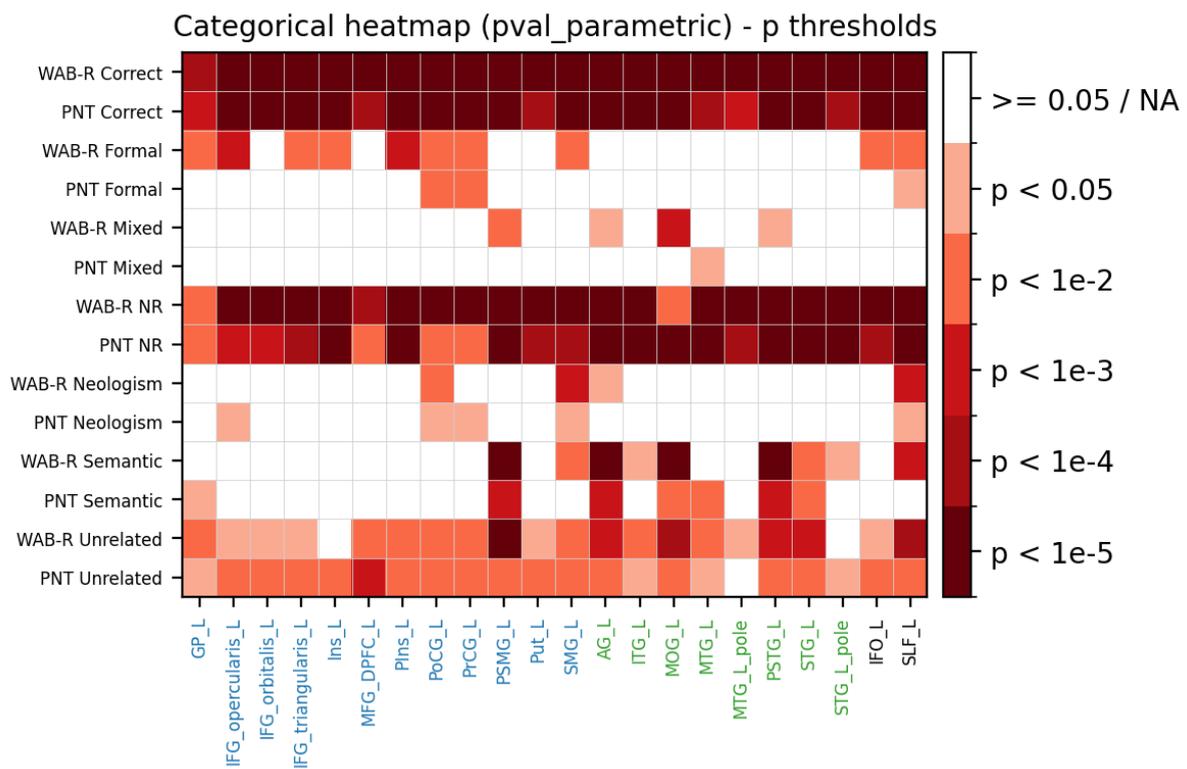

**Supplementary Figure S3. ROI-level lesion–symptom associations**. Categorical ROI-level lesion–symptom associations across PNT and WAB-R error types. Each cell reflects the statistical association between lesion load in a given JHU ROI and performance on a specific error type. Associations were obtained using mass-univariate linear regression with FDR correction (Benjamini–Hochberg). Color intensity encodes the strength of association. Behaviors are ordered to place matched PNT/WAB-R error types adjacent to each other. Regions are colored by dorsal stream (blue), ventral stream (green), and white matter tracts (black).

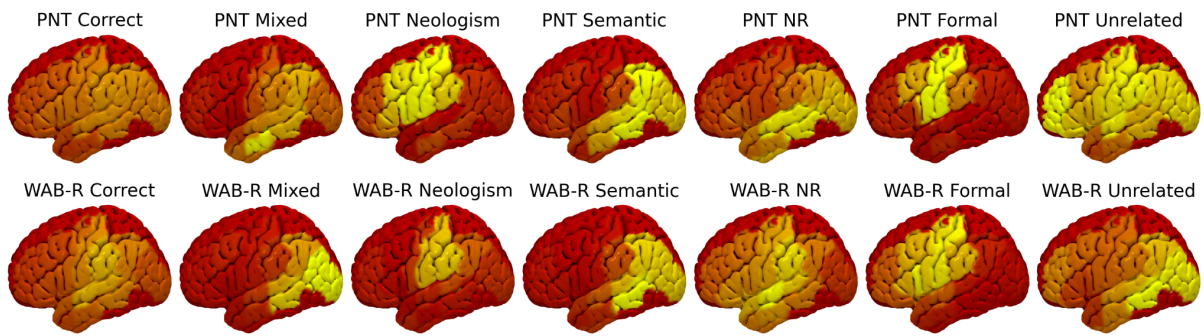

**Supplementary Figure S4. Brain renderings of lesion–symptom maps.** Lesion correlates of sentence completion and picture naming response types. Each brain rendering shows the continuous lesion–symptom association map for a specific error type. Statistical maps were transformed using -log$_{10}$(p) and rendered onto a standard cortical mesh. Colors range from red to yellow, with yellow indicating stronger lesion–symptom associations. The key result is that the 'patterns of associations', rather than the strength of the associations, are similar between specific error types measured using either the PNT or the WAB-R; colorbars are therefore omitted. Note the consistent spatial correspondence between WAB-R and PNT for correct naming, as well as for semantic and formal errors.

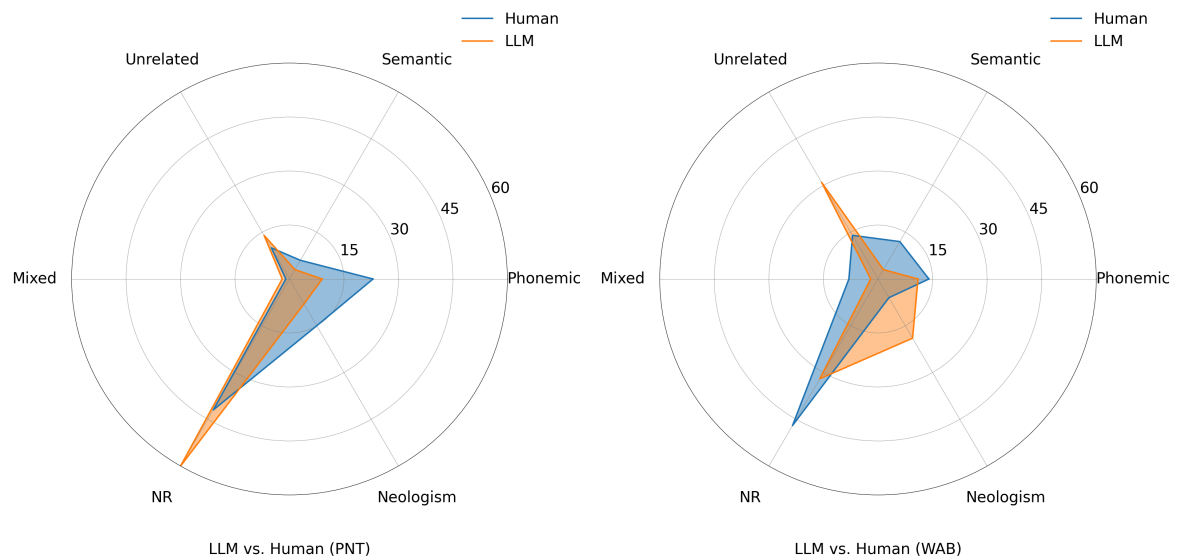

**Supplementary Figure S5. LLM error profiles vs. human error distributions.** Radar plots show the proportional distribution of naming error types produced by human participants and the large language model (LLM) across two tasks: the Philadelphia Naming Test (PNT; left) and the Western Aphasia Battery–Revised (WAB-R; right). Error categories include phonemic, semantic, unrelated, mixed, non-response (NR), and neologism errors. Human and LLM profiles are overlaid to facilitate direct comparison, demonstrating broad alignment in relative error patterns across tasks and supporting the comparability of LLM-generated errors to human aphasic naming behavior.

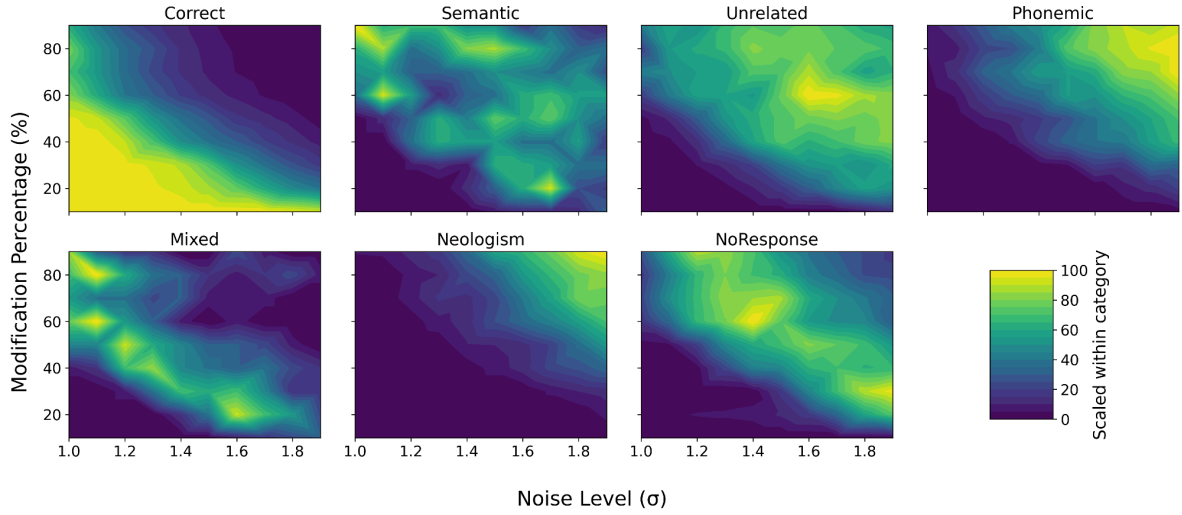

**Supplementary Figure S6. Full LLM perturbation parameter space.** Distribution of error types as a function of perturbation noise standard deviation, averaged across all transformer layers.